**Title:** High-performance automated abstract screening with large language model ensembles

**Authors:** Rohan Sanghera[1,2,*], Arun James Thirunavukarasu[1,2,*], Marc El Khoury[3,4,5], Jessica O'Logbon[6], Yuqing Chen[3], Archie Watt[7], Mustafa Mahmood[8], Hamid Butt[3], George Nishimura[3], Andrew A.S. Soltan[2,9]

**Affiliations:**

1. Medical Sciences Division, University of Oxford, Oxford, UK
2. Oxford University Hospitals NHS Foundation Trust, Oxford, UK
3. University of Cambridge School of Clinical Medicine, University of Cambridge, Cambridge, UK
4. Georgetown University School of Medicine, Georgetown University, Washington, DC, USA
5. MedStar Washington Hospital Center, Washington, DC, USA
6. GKT School of Medical Education, King's College London, London, UK
7. Oxford Medical School, Medical Sciences Divisional Office, University of Oxford, Oxford, UK
8. UCL Medical School, University College London, London, UK
9. Department of Oncology, University of Oxford, Oxford, UK

*Joint-first authors

**Correspondence details:**

Arun Thirunavukarasu: ajt205@cantab.ac.uk

**Abstract**

Large language models (LLMs) excel in tasks requiring processing and interpretation of input text. Abstract screening is a labour-intensive component of systematic review involving repetitive application of inclusion and exclusion criteria on a large volume of studies identified by a literature search. Here, LLMs (GPT-3.5 Turbo, GPT-4 Turbo, GPT-4o, Llama 3 70B, Gemini 1.5 Pro, and Claude Sonnet 3.5) were trialled on systematic reviews in a full issue of the Cochrane Library to evaluate their accuracy in zero-shot binary classification for abstract screening. Trials over a subset of 800 records identified optimal prompting strategies and demonstrated superior performance of LLMs to human researchers in terms of sensitivity ($LLM_{max}$ = 1.000, $human_{max}$ = 0.775), precision ($LLM_{max}$ = 0.927, $human_{max}$ = 0.911), and balanced accuracy ($LLM_{max}$ = 0.904, $human_{max}$ = 0.865). The best performing LLM-prompt combinations were trialled across every replicated search result (n = 119,691), and exhibited consistent sensitivity (range 0.756-1.000) but diminished precision (range 0.004-0.096). 66 LLM-human and LLM-LLM ensembles exhibited perfect sensitivity with a maximal precision of 0.458, with less observed performance drop in larger trials. Significant variation in performance was observed between reviews, highlighting the importance of domain-specific validation before deployment. LLMs may reduce the human labour cost of systematic review with maintained or improved accuracy and sensitivity. Systematic review is the foundation of evidence synthesis across academic disciplines—including evidence-based medicine—and LLMs may increase the efficiency and quality of this mode of research.

**Introduction**

Systematic review underpins evidence-based medicine (EBM) as the primary method for synthesising data from previously reported clinical studies[1,2] as well as knowledge from research in non-medical fields.[1,3,4] Good practices include transparent reporting and reproducible methodology, and checklists and guidance exist to support adherence to accepted standards of conduct and reporting.[1,5] Some tasks involved in the systematic review process can be labour-intensive, repetitive, and text-based with formulaic and algorithmic schema used to maximise reproducibility.[6] Examples include trialling search strategies, screening abstracts and full texts for inclusion, and extracting data from included studies.[3,7]

Abstract screening is the process of selecting articles identified by the search strategy that meet pre-specified criteria for inclusion, and is typically performed by two or more researchers with domain-specific expertise. Screeners use the title and abstract of each record to determine eligibility and make decisions to include or exclude accordingly. Tools to streamline abstract screening are already in wide use—such as hosted on Rayyan and Covidence—but researchers using these tools are limited to a maximum rate of screening of around 100 abstracts per minute.[8] Use of emerging artificial intelligence (AI) applications has been posited as a means of improving the accuracy and efficiency of abstract screening.[9,10]

Computational natural language processing has advanced significantly with the development and deployment of large language models (LLMs).[11] LLMs are pretrained on large volumes of human-produced text and then instruction-tuned on a wide variety of tasks to develop remarkable abilities to interpret and generate text in multiple languages.[11] In medicine, LLMs have garnered significant attention for attaining comparable results to clinicians in examinations and other reasoning tasks, but are yet to be deployed in a decision-making capacity in real-world settings.[12,13] Healthcare research offers an arena in which LLMs may be deployed with less direct risk to patients, and automating systematic review is one such avenue of research.[14] However, high accuracy is critical to ensure that the conclusions drawn are valid, as systematic reviews are the highest weighted evidence when designing treatment algorithms and providing advice to clinicians and patients.[7,15] As a reasoning and binary classification task, abstract screening is may be amenable to automation using LLMs.

We aimed to provide a general estimate of the abstract screening performance of a variety of LLMs; demonstrate an effective workflow to optimise accuracy and sensitivity of automated screening; and show how different LLMs and human researchers may be combined to maximise efficiency and accuracy. We approached abstract screening as a zero-shot binary classification problem, thereby maximising generalisability to other systematic reviews and literature syntheses by limiting the requirement for domain-specific fine-tuning or prompt engineering.

**Methods**

*Development of the LLM pipeline for automated abstract screening*

LLM screening was undertaken via an application programming interface (API) using Jupyter notebooks. GPT-3.5 Turbo (gpt-3.5-turbo-0125), GPT-4 Turbo (gpt-4-0125-preview), and GPT-4o (gpt-4o-2024-05-13) models were accessed through the Azure OpenAI Service, using the openai (1.23.2) Python package. Llama 3 70B (meta-llama-3-70b) was hosted on Replicate.com and accessed using the replicate

(0.26.0) Python package. Claude Sonnet 3.5 (claude-3-5-sonnet@20240620) and Gemini 1.5 Pro (gemini-1.5-pro-001) models were accessed through Google Cloud's Vertex AI platform, using the google-cloud-aiplatform (1.62.0) Python package.

Multiple prompts were developed to investigate the effect of prompt engineering on abstract screening performance. These prompts were developed through an iterative process based on small scale experiments with fewer than 100 records screened at a time, using the GPT-3.5 API. In addition to a control prompt where only the title of each record was presented, a range of prompts were developed where the title and abstract were presented with instructions containing variable levels of bias towards inclusion: 'none', 'mild', 'moderate', 'heavy', and 'extreme'. The prompt for Llama 3 was subtly adjusted to incorporate special tokens and align with its specific prompt structure. Enforcing chain of thought reasoning and numerical scoring were trialled but without yielding improved sensitivity or accuracy and were therefore not incorporated into the final prompts. The final wording of each used prompt is provided in Supplementary Material 1.

The inclusion criteria and exclusion criteria for each systematic review were converted into numbered lists, and, together with the titles of each review, were parsed into Python dictionaries. The model prompt was iteratively updated for each record in a Pandas (2.1.1) DataFrame, using the respective parent review's dictionaries and the article's title and abstract. Based on insights gained from preliminary trials, the 'temperature' and 'max_tokens' parameters were specified as 0.2 and 5, respectively, to produce a more deterministic output of limited length which was best suited to abstract screening. Other adjusted parameters included

'top p', 'frequency_penalty', and 'presence_penalty'. However, adjustment of these parameters did not yield significant improvements in performance or reliability, so their default values were retained for subsequent deployment. In cases where the API returned an error or an invalid decision, the query was repeated, with exponential backoff applied to address rate-liming.

In some cases, models returned an error message, content violation note, or no interpretable output, despite repeating queries. These were listed as 'include' for analysis for the same reason as prompt bias being designed to tend towards inclusion: to avoid excluding potentially eligible records before full text screening. This schema aimed to maximise model sensitivity even at the expense of overall accuracy, as false negatives (exclusion of eligible records) are a more damaging error than false positives (including ineligible records for full text screening), because eligible records are permanently lost and therefore not used in the evidence synthesis.

*Validation of the LLM application*

All systematic reviews from the latest issue of the Cochrane Database of Systematic Reviews at the time of protocol development (2023, Issue 8) were used for experiments.[16–38] These were selected due to the reputation of the Cochrane Library for gold-standard methodology, consistency in reporting (including search strategies and inclusion criteria), and unbiased coverage of topics across medicine and surgery. The search strategy from each review was replicated using the same keyword combinations on databases as listed in reviews' appendices. Records published after the date of search listed in the reviews were excluded from subsequent analyses. The inclusion lists of each review were used as the ground-truth: gold standard examples of articles which should have been included on the basis of the reviews' protocols.

Replicated and de-duplicated inclusion lists and searches for each of the 23 systematic reviews were imported in RIS file format, using the rispy (0.81) python package, and parsed into DataFrames. During data cleaning, 8,604 articles with missing abstracts were excluded from further analysis, constituting 6.71% of the total data set. A total of 119,695 articles were eligible for analysis and experiments. A subset of 800 articles was generated using the inclusion DataFrame and a random, seeded sample of 23 excludes from each review, to create a dataset with improved balance for smaller scale experiments.

Search results for each review were used as inputs to the LLM application, which used inclusion and exclusion criteria as detailed in the review protocol to decide whether to include or exclude each study. Replicated searches often returned different numbers of records to those reported in the original reviews, likely due to a combination of use of sources other than databases, as well as errors in search strategy reporting.[39] Records published after the original reviews were removed from experiments to avoid conflicts between genuine eligibility based on criteria and lack of representation in the review. Confusion matrices were constructed using the total numbers of studies identified in replicated searches, using the following definitions:

- True positive: eligible record correctly included
- True negative: ineligible record correct excluded
- False positive: ineligible record incorrectly included
- False negative: eligible record incorrectly excluded

Sensitivity, specificity, and accuracy calculations were subsequently performed to provide interpretable measures of abstract screening performance. For one review where no eligible records were included, sensitivity was represented as 100% rather than indeterminate to facilitate quantitative comparisons.[35]

*Comparative experiments*

All LLMs were trialled using every designed prompt across a subset of 800 records. All records included in each Cochrane review were used, with randomly selected records from the replicated search (excluded in the Cochrane reviews) used for experiments. To establish consistency, LLM trials were repeated across the whole subset of records. Three human researchers undertook screening using original review criteria and no other prompts, across the same subset of records, with confusion matrices and performance metrics calculated in the same way as for LLMs. Kappa statistics were calculated for each LLM and human combination of results, including repeat trials, to gauge relative consistency. Correlation analysis was undertaken to explore whether human and LLM performance was aligned with respect to the reviews. To quantify performance, precision (positive predictive value) and recall (sensitivity) were used as the primary outcome variables of interest, as these provide more informative comparisons than sensitivity and specificity for binary classifiers where 'include' decisions are relatively rare.[40]

For each LLM, one prompt was used for further experiments across the entire set of records returned by replicated searches for each review. Prompt selection was based on optimal performance in terms of balanced accuracy, used instead of raw accuracy due to the significant imbalance in the dataset due to relatively few included articles. However, for GPT-3.5 the 'heavy' bias prompt was used despite results for the prompts with 'mild' bias or 'none' exhibiting higher balanced accuracy, because perfect sensitivity was exhibited with the heavy prompt which would be more compatible with autonomous deployment. GPT-4 was not trialled across the whole dataset, as its updated version, GPT-4o, exhibited better performance, efficiency, and cost. Qualitative comparisons were made to the specificity of the original authors of each systematic review based on their reported numbers of records excluded after abstract screening and records ultimately included; sensitivity was 100% for the original authors in every case because their decisions were used to determine which records were supposed to be included.

All experiments were conducted in Python (Python Software Foundation, Wilmington, Delaware, USA; version 3.11.5). Data analysis and visualisation were conducted in R (R Foundation for Statistical Computing, Vienna, Austria; version 4.2.1) and Affinity Designer (Serif Europe Ltd., West Bridgford, UK; version 1.10.6). All code required to replicate experiments and analysis is hosted on GitHub (https://github.com/RohanSanghera/GEN-SYS).

**Results**

*Prompt wording determines LLM abstract screening performance*

23 reviews were used for experiments, comprising the entirety of 2023 Issue 8 of the *Cochrane Database of Systematic Reviews* (Table 1).[16–38] These reviews exhibited a wide range of specialties, interventions, and sizes in terms of included studies and participants. Two reviews features lead authors with the same name, and were referred to as Singh-1[33] and Singh-2[32] to distinguish between them.

| Lead author | Title | n(search results) | n(included studies) |
|---|---|---|---|
| Bellon | Perioperative glycaemic control for people with diabetes undergoing surgery | 3693 | 23 |
| Buchan | Medically assisted hydration for adults receiving palliative care | 5043 | 4 |
| Clezar | Pharmacological interventions for asymptomatic carotid stenosis | 6476 | 31 |
| Cutting | Intracytoplasmic sperm injection versus conventional in vitro fertilisation in couples with males presenting with normal total sperm count and motility | 3092 | 3 |
| Dopper | High flow nasal cannula for respiratory support in term infants | 8768 | 8 |
| Ghoraba | Pars plana vitrectomy with internal limiting membrane flap versus pars plana vitrectomy with conventional internal limiting membrane peeling for large macular hole | 2690 | 5 |
| Hjetland | Vocabulary interventions for second language (L2) learners up to six years of age | 6238 | 12 |
| Karkou | Dance movement therapy for dementia | 706 | 3 |
| Lin | Hyperbaric oxygen therapy for late radiation tissue injury | 773 | 16 |
| Lynch | Interventions for the uptake of evidence-based recommendations in acute stroke settings | 20319 | 7 |
| Malik | Fibrin-based haemostatic agents for reducing blood loss in adult liver resection | 3685 | 20 |
| Mohamed | Prostaglandins for adult liver transplanted recipients | 304 | 11 |
| Roy | Interventions for chronic kidney disease in people with sickle cell disease | 5891 | 28 |
| Santos | Prophylactic anticoagulants for non-hospitalised people with COVID-19 | 17380 | 5 |
| Setthawong | Extracorporeal shock wave lithotripsy (ESWL) versus percutaneous nephrolithotomy (PCNL) or retrograde intrarenal surgery (RIRS) for kidney stones | 1880 | 21 |
| Sévaux | Paracetamol (acetaminophen) or non-steroidal anti-inflammatory drugs, alone or combined, for pain relief in acute otitis media in children | 10826 | 4 |
| Singh-1 | Blue-light filtering spectacle lenses for visual performance, sleep, and macular health in adults | 195 | 17 |
| Singh-2 | Interventions for bullous pemphigoid | 312 | 16 |
| Sulewski | Topical ophthalmic anesthetics for corneal abrasions | 7016 | 9 |
| Sulistyo | Enteral tube feeding for amyotrophic lateral sclerosis/motor neuron disease | 189 | 0 |
| White | Oxygenation during the apnoeic phase preceding intubation in adults in prehospital, emergency department, intensive care and operating theatre environments | 13549 | 22 |
| Younis | Hydrogel dressings for donor sites of split-thickness skin grafts | 425 | 2 |
| Zhu | Expanded polytetrafluoroethylene (ePTFE)-covered stents versus bare stents for transjugular intrahepatic portosystemic shunt in people with liver cirrhosis | 246 | 4 |

**Table 1**: Characteristics of 23 systematic reviews taken from 2023 Issue 7 of the Cochrane Database of Systematic Reviews, used for all experiments. The reviews covered a broad range of specialties, interventions, sample sizes (in terms of studies and participants), and methodologies (meta-analyses and narrative syntheses). Numbers of records and included studies are based on replicated searches undertaken for experimental purposes, rather than the numbers reported in the original reviews.

Across the subset of 800 records, LLM performance varied remarkably when different prompts were used (Figure 1). As sensitivity increased, precision tended to decrease (Pearson's correlation coefficient, R = -0.47, 95% confidence interval -0.53 to -0.42, *p < 0.001)*, in-keeping with a lower threshold for inclusion. Recall varied significantly as the prompt was changed, consistent with prompt wording being responsible for the difference in threshold for inclusion (Kruskal-Wallis test, $χ^2$ = 62.5, *p* < 0.001). Moreover, differences were directed in the same manner as the language used in the prompt: a heavier bias towards inclusion resulted in more records being included.

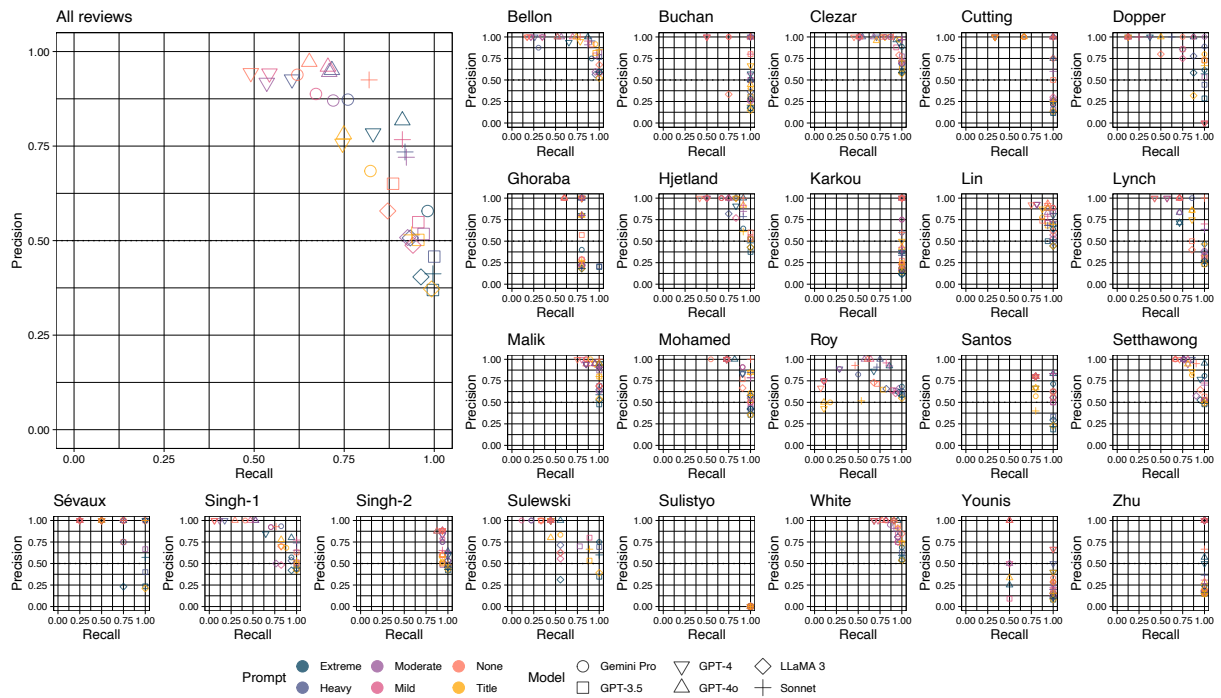

**Figure 1**: Precision (positive predictive value) and recall (sensitivity) of six LLMs tasked with automated abstract screening across a range of six prompts with different information content and bias towards inclusion. Sensitivity was deemed 100% for all models working with Sulistyo et al, 2023, as there were no articles deemed eligible for inclusion in the original review. Performance was highly variable between models and across different prompts but was comparable to human researchers conducting the same abstract screening task. When used with a prompt containing a 'heavy' bias towards inclusion, GPT-3.5 exhibited perfect (100%) sensitivity across every review, meaning that all eligible articles were correctly included. For other models, the optimal prompt taken forward in further experiments was determined by the highest calculated balanced accuracy: 'none' (no bias towards inclusion) for Llama 3 and Sonnet, 'heavy' bias for Gemini Pro, and 'extreme' bias for GPT-4o. Balanced accuracy was optimal for GPT-4 with the 'extreme' bias prompt, but was inferior to its successor model, GPT-4o, in addition to exhibiting worse efficiency.

*LLM accuracy and consistency relative to human researchers drops with sample size*

The performance of human researchers replicating abstract screening undertaken in each Cochrane review lay within the range of accuracy, sensitivity, and specificity of LLMs tasked with screening the same abstracts (Table 2). For every calculated performance metric, an LLM (either GPT-3.5, GPT-4, or Sonnet) exhibited the highest score; higher than all three human researchers. Similar comparative performance was observed when results were stratified by review (Supplementary Material 2).

| Human/Model | Sensitivity (Recall) | Specificity | Balanced Accuracy | Precision (PPV) | NPV | F1-score |
|---|---|---|---|---|---|---|
| Alpha | 0.745 | 0.962 | 0.854 | 0.910 | 0.881 | 0.819 |
| Bravo | 0.720 | 0.964 | 0.842 | 0.911 | 0.870 | 0.804 |
| Charlie | 0.775 | 0.955 | 0.865 | 0.897 | 0.892 | 0.832 |
| GPT-3.5 | 1.000 | 0.393 | 0.697 | 0.458 | 1.000 | 0.628 |
| GPT-4 | 0.605 | 0.975 | 0.857 | 0.927 | 0.828 | 0.732 |
| GPT-4o | 0.911 | 0.896 | 0.904 | 0.818 | 0.952 | 0.862 |
| Gemini 1.5 Pro | 0.760 | 0.943 | 0.852 | 0.873 | 0.885 | 0.813 |
| LLaMA 3 | 0.871 | 0.675 | 0.773 | 0.578 | 0.911 | 0.695 |
| Sonnet 3.5 | 0.819 | 0.966 | 0.893 | 0.925 | 0.913 | 0.869 |

**Table 2**: Performance of three human researchers (Alpha, Bravo, and Charlie) and six LLMs (GPT-3.5 Turbo, GPT-4 Turbo, GPT-4o, Gemini 1.5 Pro, LLaMA 3 70B, and Claude Sonnet 3.5; all used with their respective optimal prompts) replicating abstract screening for a subset of 800 records obtained by replicating Cochrane review search strategies and inclusion lists. LLMs (specifically GPT-3.5, GPT-4, and Sonnet) exhibited the highest performance in terms of every measured metric.

The consistency of LLM decisions, quantified by calculating Kappa statistics, varied between reviews when abstract screening was repeated (Table S1). Where the optimal prompt was used, $\kappa_{GPT-3.5}$ ranged between 0.487-1.000 (median = 0.868), $\kappa_{GPT-4}$ between 0.870-1.000 (median = 0.957), $\kappa_{GPT-4o}$ between 0.787-1.000 (median = 0.941), $\kappa_{Gemini\ Pro}$ between 0.927-1.000 (median = 1.000), $\kappa_{Llama\ 3}$ between 0.642-1.000 (median = 0.881), and $\kappa_{Sonnet\ 3.5}$ between 0.903-1.000 (median = 1.000). Human researchers screening the same abstracts exhibited more inconsistency than the LLMs but similar variation across reviews, with a median Kappa statistic of 0.827 (range -0.045 to 1.000).

Correlational analysis was undertaken to explore whether review-centric factors were likely to determine observed variation in agreement and performance across different systematic reviews. Despite noise generated by dependency of performance on the model or human evaluator, and prompt used with the LLMs, a consistent association between human and LLM performance was observed across the Cochrane reviews (Figure 2). Consistent association between human performance and LLM performance was determined by calculation of the coefficient of determination. Positive association was greatest for sensitivity ($R^2 = 0.196$), balanced accuracy ($R^2 = -.193$), F1-score ($R^2 = 0.193$), and positive predictive value ($R^2 = 0.299$); and lower for negative predictive value ($R^2 = 0.073$) and specificity ($R^2 = 0.012$). These results indicated that review-centric factors—perhaps clarity and comprehensiveness of reporting—affected abstract screening performance in addition to LLM and human factors such as intrinsic aptitude, expertise, and prompt engineering.

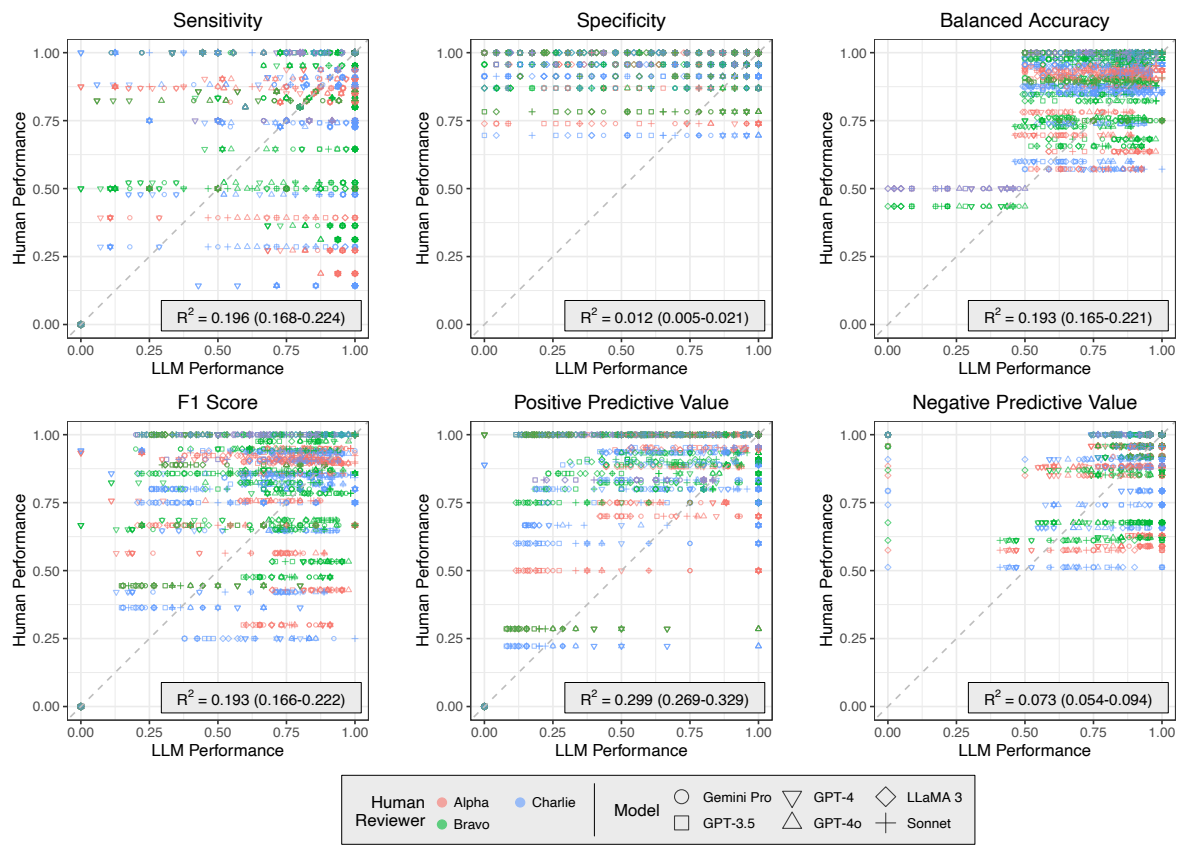

**Figure 2**: Correlational analysis undertaken to investigate whether review-centric factors influenced the abstract screening performance of LLMs and human researchers replicating the work of Cochrane systematic review authors. Considerable noise is observed, due mostly to established variation with respect to human researcher, LLM, and prompt wording. However, despite this noise, between 1.2% and 29.9% of the variation in human performance was predictable based on LLM performance, with higher coefficients of determination for sensitivity, F1-score, balanced accuracy, and positive predictive value. It is likely that review-centric factors—such as clarity and comprehensiveness of reporting—contribute to this relationship. Less association was observed for sensitivity and negative predictive value, perhaps in part due to less overall variation in human and LLM performance when measured with those metrics.

LLM precision was expected to drop when tested over every single replicated search result, as all additional records were ineligible for inclusion. To provide a comparator representing the expected performance ceiling, original numbers of records according to the Cochrane reviews were used to calculate performance metrics.[16–38] The drop in LLM precision relative to that in smaller scale experiments was remarkable, and was below that of the original Cochrane reviewers (Table 3). However, as sensitivity was preserved, models appeared feasible of saving researcher time and effort by identifying ineligible records automatically.

| Human/Model | Sensitivity (Recall) | Specificity | Balanced Accuracy | Precision (PPV) | NPV | F1-score |
|---|---|---|---|---|---|---|
| Cochrane | 1.000 | 0.993 | 0.996 | 0.235 | 1.000 | 0.381 |
| GPT-3.5 | 1.000 | 0.419 | 0.710 | 0.004 | 1.000 | 0.008 |
| GPT-4o | 0.904 | 0.949 | 0.926 | 0.038 | 1.000 | 0.074 |
| Gemini 1.5 Pro | 0.756 | 0.976 | 0.866 | 0.068 | 0.999 | 0.125 |
| LLaMA 3 | 0.841 | 0.776 | 0.809 | 0.008 | 1.000 | 0.017 |
| Sonnet 3.5 | 0.823 | 0.982 | 0.903 | 0.096 | 1.000 | 0.172 |

**Table 3**: Performance of five LLMs used with their respective optimal prompts across every record returned by replicated searches, compared to an expected performance ceiling calculated from the reported number of included abstracts in the original Cochrane systematic reviews used for experiments. For each review, the search strategy was replicated and implemented to use a common (fair) number of records screened between LLM trials and human benchmark. In contrast to comparisons with human researchers replicating screening, performance calculated from Cochrane review authors' reported numbers was superior to all tested LLMs. Precision was remarkably low for all screeners, likely driven by a low prevalence of eligible articles for inclusion.

*Ensemble configurations approach gold-standard abstract screening performance*

LLMs and human researcher decisions were combined in series and in parallel, in six distinct configurations (Figure 3). Combination in series meant that both component decisions had to be 'include' for articles to be included; combination in parallel meant that only one 'include' decision was required for inclusion. As predicted, series ensembles exhibited greater average precision as the criteria for inclusion were stricter; while parallel ensembles exhibited greater sensitivity as they had a lower barrier to inclusion. 66 ensembles belonging to two parallel configuration schema exhibited perfect sensitivity: LLM and human in parallel, and LLM and LLM in parallel. Every ensemble with perfect sensitivity had a higher precision than recorded by the original Cochrane reviewers (Figure 3). Many ensembles comprised of two LLMs in series approached perfect sensitivity, the closest being GPT-3.5 with heavy bias prompt and Sonnet with extreme bias prompt (sensitivity = 0.996).

The highest precision of an LLM-LLM ensemble with perfect sensitivity was 0.458, exhibited by GPT-3.5 with heavy bias prompt combined with any of the following models in parallel: Sonnet with no bias prompt, GPT-4 with no bias prompt, GPT-4 with mild bias prompt, GPT-4 with moderate bias prompt, GPT-4 with heavy bias prompt, GPT-4o with no bias prompt, GPT-4o with mild bias prompt, GPT-4o with moderate bias prompt, GPT-4o with heavy bias prompt, and Gemini Pro with no bias prompt. One LLM-human ensemble attained the same precision of 0.458 with perfect sensitivity: GPT-3.5 with heavy bias and Bravo. Six LLM-human ensembles and 60 LLM-LLM ensembles attained perfect sensitivity in total.

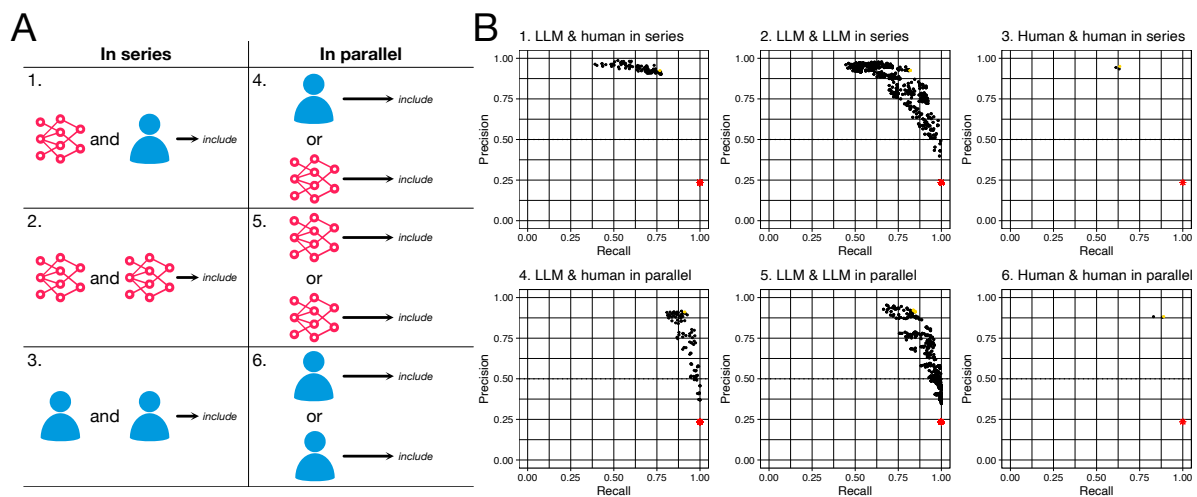

**Figure 3**: (A) Schematics describing six distinct configurations for incorporation of LLM and human

decisions into a binary ensemble system. (B) Precision (positive predictive value) and recall (sensitivity) of every ensemble permutation combining LLMs, prompts, and human researchers in each configuration (three in series and three in parallel). For each configuration, the ensemble with the highest calculated accuracy is coloured gold. For comparison, calculated performance from the original Cochrane reviews is indicated with red asterisks. Maximal sensitivity (recall) is necessary to justify deployment such that eligible articles are not erroneously excluded by the ensemble system. 66 ensembles across two parallel configurations obtained maximal sensitivity, all with higher precision than the Cochrane reviewers, albeit across a much lower number of records. The highest precision associated with maximal sensitivity for each configuration was 0.458, exhibited by GPT-3.5 with heavy bias prompt when combined with one of ten other model-prompt combinations. While many LLM-LLM in series ensembles approached perfect sensitivity, the best performing system was GPT-3.5 with heavy bias prompt and Sonnet with extreme bias prompt (sensitivity = 0.996).

However, while ensemble performance exceeded the statistical benchmark set by the original reviewers, caution is required as the original reviewers screened a far greater number of abstracts. As seen with results from individual LLMs, precision tends to drop as the number of ineligible articles to screen is increased. LLM-only ensembles exhibited a similar drop in precision (range = 0.003-0.139) when tested across all available records (Table S1), although many ensembles exhibited higher precision than individual LLMs. LLM-human ensembles could not be tested on more records than were screened by researchers.

**Discussion**

Optimal combinations of LLMs and prompts exhibited perfect sensitivity (recall), and comparable precision (positive predictive value) to researchers replicating the abstract screening undertaken in gold standard systematic reviews. However, LLM precision dropped significantly when tested at larger scale, primarily because relatively few screened articles are eligible for inclusion. Human precision also exhibited a drop when screening a greater number of abstracts, even where researchers screening greater numbers of abstracts had more domain-specific expertise than non-expert researchers screening a subset of abstracts. This is because abstract screening is intended to maximise sensitivity at the expense of precision, as it is imperative that eligible records are not erroneously excluded; subsequent full-text screening determines which articles to finally include. Therefore, despite potentially exhibiting lower precision than humans with maximal domain-specific expertise, LLMs—particularly in tandem with human researchers—can be of great assistance with abstract screening. Moreover, LLM abstract screening may improve the quality of evidence synthesis undertaken by researchers without domain-specific expertise by reducing the number of eligible records that are lost.

The relative performance of LLMs may be more favourable than comparisons to the original reviews suggest. The performance ceiling calculated from the original reviews was likely inflated by use of review authors' decisions to define the ground truth, authors' subject matter expertise and preconceived notions of what types of study were supposed to be included, as well as mistakes or omissions in descriptions of the search and screening process. When formally tested, trained human researchers exhibit a lower abstract screening sensitivity than the performance ceiling calculated here from Cochrane review data: 87%, improving to 97% when two human researchers screen each abstract.[41] LLMs exceeded this benchmark, and may therefore complement conventional screening and reduce the workload for human researchers. However, domain-specific validation is necessary to demonstrate the sensitivity is high enough to justify autonomous deployment, as performance varies significantly between reviews. In addition, clarity of the review question and inclusion criteria is critical to facilitate reproducible LLM screening—variable replication results from LLMs and human researchers suggest that the quality of reporting is often insufficient, agreeing with many previous studies appraising systematic reviews.[6,42]

Previous proof-of-concept studies have evaluated LLM abstract screening but have been highly restricted in terms of subject matter or failed to provide fair comparators to contextualise results.[9,43–45] Here, a prompt engineering strategy for automated abstract screening worked well with a wide variety of LLMs, albeit with variable performance between reviews. Potential applications could change the methodology of systematic review by working in series or in parallel with human researchers. With sufficient sensitivity demonstrated over a subset of studies, models could be entrusted with autonomously pre-screening studies to reduce the number of studies requiring human evaluation: working in series to maximise efficiency of screening but accepting that erroneously excluded eligible studies cannot be salvaged. Automated systems may also be used in place of a second reviewer in parallel with human researchers, halving the initial screening workload by operating across the whole number of identified records, and potentially capturing mistakenly excluded and included studies to improve screening accuracy and reduce the burden of full-text screening. For models designed to work in parallel, sensitivity may be sacrificed to maximise accuracy and thereby efficiency as more ineligible records can be excluded before full text screening. Specific fine-tuning may be employed to optimise performance and model behaviour, although careful validation is required as customised models do not necessarily exhibit superior performance.[46] Alternatively, rather than binary output to determine whether records should be included or excluded, a combination of prompt engineering and fine-tuning could be employed to generate uncertainty estimates which could guide human researchers to review records where model decisions are less likely to be accurate.[47]

Three limitations may have affected the study's results and conclusions. First, representativeness was limited although a full issue of the *Cochrane Database of Systematic Reviews* was used to test across a broad range of medical topics. Inter-review variation shows that automated screening may be better suited to some subjects than others, and applications should be specifically validated within a subject or topic if used. Further work may seek to explore where LLM screening is most effective, and how review protocols and screening criteria could be better designed to facilitate automation. LLMs could even be used as a tool to quantify the clarity and reproducibility of screening described in systematic review reports. Second, the study may have exhibited an optimisation bias in favour of GPT-3.5 as initial prompt engineering was undertaken using that LLM in smaller scale experiments. This was due to relative ease and lower cost of access. Further improvement in the performance of each LLM is likely feasible with more intensive prompt engineering, which could be specifically directed to the aims of a single review.[11] Finally, the performance of Cochrane reviewers were likely inflated by their use as both a comparator and as ground truth, as well as due to any mistakes, omissions, or unclearly communicated aspects in the reviewers' search and screening strategies.[39] While the performance of the original reviewers serves as a useful benchmark corresponding to maximal possible accuracy, the alternative comparator provided by independent researchers replicating screening is a more useful gauge of the relative strengths and limitations of LLM-based screening, and also lay closer to previous estimates of human screening performance.[41]

Further work is required to integrate automated abstract screening into the conventional workflow of systematic review: our approach requires accessing an API with a spreadsheet containing details from every study identified at the search stage. By providing comprehensive detail about the LLMs used and our broader methodology, we aim to maximise reproducibility of our results and access to automated abstract screening.[48] However, code-free solutions would enable more researchers to leverage automated abstract screening in their research.[49] The institution of EBM relies heavily upon accurate syntheses of available evidence to answer clinical questions, of which systematic review forms a critical component. It is therefore critical that implementation of automated abstract screening does not compromise the quality or reproducibility of systematic review.[10] We would recommend authors report any use of automated screening technology clearly enough for other researchers to replicate their approach, including details about the model and prompt used, and how automated screening contributed to inclusion decisions. Ideally, the Preferred Reporting Items for Systematic Reviews and Meta-Analyses (PRISMA) should include these details as automated screening becomes common practice.[5]

**Conclusion**

LLMs can facilitate automated abstract screening with high sensitivity, best operating in parallel with human researchers. Automated abstract screening may improve the efficiency and quality of systematic review and could thereby improve the practice of EBM. LLM performance is subject-specific but can be optimised through prompt engineering, and researchers are advised to conduct domain-specific validation before unsupervised deployment. Through more accurate, efficient, and thorough systematic review of clinical questions, clinicians and researchers can ensure that more patients receive optimal care.

**Financial declaration**: This study was supported by the HealthSense Research Fund, and the Microsoft Research Accelerating Foundation Models Research Award. The funders had no input in to the design, conduct, and reporting of the study.

**Competing interests:** The authors have no intellectual, financial, or personal competing interests to disclose.

**Legends**

*Figure 1*: Precision (positive predictive value) and recall (sensitivity) of six LLMs tasked with automated abstract screening across a range of six prompts with different information content and bias towards inclusion. Sensitivity was deemed 100% for all models working with Sulistyo et al, 2023, as there were no articles deemed eligible for inclusion in the original review. Performance was highly variable between models and across different prompts but was comparable to human researchers conducting the same abstract screening task. When used with a prompt containing a 'heavy' bias towards inclusion, GPT-3.5 exhibited perfect (100%) sensitivity across every review, meaning that all eligible articles were correctly included. For other models, the optimal prompt taken forward in further experiments was determined by the highest calculated balanced accuracy: 'none' (no bias towards inclusion) for Llama 3 and Sonnet, 'heavy' bias for Gemini Pro, and 'extreme' bias for GPT-4o. Balanced accuracy was optimal for GPT-4 with the 'extreme' bias prompt, but was inferior to its successor model, GPT-4o, in addition to exhibiting worse efficiency.

*Figure 2*: Correlational analysis undertaken to investigate whether review-centric factors influenced the abstract screening performance of LLMs and human researchers replicating the work of Cochrane systematic review authors. Considerable noise is observed, due mostly to established variation with respect to human researcher, LLM, and prompt wording. However, despite this noise, between 1.2% and 29.9% of the variation in human performance was predictable based on LLM performance, with higher coefficients of determination for sensitivity, F1-score, balanced accuracy, and positive predictive value. It is likely that review-centric factors—such as clarity and comprehensiveness of reporting—contribute to this relationship. Less association was observed for sensitivity and negative predictive

value, perhaps in part due to less overall variation in human and LLM performance when measured with those metrics.

*Figure 3*: (A) Schematics describing six distinct configurations for incorporation of LLM and human decisions into a binary ensemble system. (B) Precision (positive predictive value) and recall (sensitivity) of every ensemble permutation combining LLMs, prompts, and human researchers in each configuration (three in series and three in parallel). For each configuration, the ensemble with the highest calculated accuracy is coloured gold. For comparison, calculated performance from the original Cochrane reviews is indicated with red asterisks. Maximal sensitivity (recall) is necessary to justify deployment such that eligible articles are not erroneously excluded by the ensemble system. 66 ensembles across two parallel configurations obtained maximal sensitivity, all with higher precision than the Cochrane reviewers, albeit across a much lower number of records. The highest precision associated with maximal sensitivity for each configuration was 0.458, exhibited by GPT-3.5 with heavy bias prompt when combined with one of ten other model-prompt combinations. While many LLM-LLM in series ensembles approached perfect sensitivity, the best performing system was GPT-3.5 with heavy bias prompt and Sonnet with extreme bias prompt (sensitivity = 0.996).

*Table 1*: Characteristics of 23 systematic reviews taken from 2023 Issue 7 of the Cochrane Database of Systematic Reviews, used for all experiments. The reviews covered a broad range of specialties, interventions, sample sizes (in terms of studies and participants), and methodologies (meta-analyses and narrative syntheses). Numbers of records and included studies are based on replicated

searches undertaken for experimental purposes, rather than the numbers reported in the original reviews.

*Table 2*: Performance of three human researchers (Alpha, Bravo, and Charlie) and six LLMs (GPT-3.5 Turbo, GPT-4 Turbo, GPT-4o, Gemini 1.5 Pro, LLaMA 3 70B, and Claude Sonnet 3.5; all used with their respective optimal prompts) replicating abstract screening for a subset of 800 records obtained by replicating Cochrane review search strategies and inclusion lists. LLMs (specifically GPT-3.5, GPT-4, and Sonnet) exhibited the highest performance in terms of every measured metric.

*Table 3*: Performance of five LLMs used with their respective optimal prompts across every record returned by replicated searches, compared to an expected performance ceiling calculated from the reported number of included abstracts in the original Cochrane systematic reviews used for experiments. For each review, the search strategy was replicated and implemented to use a common (fair) number of records screened between LLM trials and human benchmark. In contrast to comparisons with human researchers replicating screening, performance calculated from Cochrane review authors' reported numbers was superior to all tested LLMs. Precision was remarkably low for all screeners, likely driven by a low prevalence of eligible articles for inclusion.

*Table S1*: Kappa statistics quantifying the consistency of LLM and human researcher decisions, for abstract screening performance across all Cochrane reviews as well as stratified by systematic review. Consistency varied across reviews, with similar patterns apparent in human researcher results.

*Table S2*: Performance of LLM-LLM ensembles tested across every record returned by replicated searches for each Cochrane review. Precision is lower than recorded

in shorter experiments, as expected, but is superior to individual LLMs (and closer to the performance ceiling calculated from numbers reported in the Cochrane reviews).

*Supplementary Material 1*: Prompts used in experimental development of the large language model (LLM) application. A range of prompts were initially trialled using GPT-3.5, and the best-performing prompts were used across the entire body of search results from each review with a range of LLMs. More sophisticated prompt engineering strategies such as enforced chain-of-thought reasoning and quantification of output did not yield improvements in performance, so were not taken forwards. All trials and experiments were zero-shot in that no domain-specific fine-tuning was undertaken, and prompts did not contain examples of correct inclusion or exclusion decisions. This maximised the generalisability of results beyond the reviews involved in experiments here, and minimised cost and technical complexity, maximising breadth of access to automated abstract screening.

*Supplementary Material 2*: Full results of experiments trialling LLMs and humans across a subset of records, and LLMs across all records from replicated searches reported by 23 Cochrane reviews. Significant variation in performance between reviews was observed, but comparative performance between LLMs, prompts, and human researchers was broadly consistent.